\pdfoutput=1
\documentclass{article}

\usepackage{microtype}
\usepackage{graphicx} 
\usepackage{bmpsize}
\usepackage{cuted}% for environment `strip`
\usepackage{subfigure}
\usepackage{lipsum}
\usepackage{amssymb,amsmath,amsthm}
\usepackage{mathtools}
\usepackage{amsfonts}
\usepackage{booktabs} % for professional tables
\usepackage[noend]{algpseudocode}
\usepackage{accents}
\usepackage{enumitem}
\usepackage{booktabs,caption}
\usepackage{threeparttable}
\usepackage{multirow} % Required for multirows

\theoremstyle{definition}

\usepackage[accepted]{icml2019}

\renewcommand{\sec}[1]{Section~\ref{#1}}
\newcommand{\fig}[1]{Figure~\ref{#1}}
\newcommand{\eq}[1]{Equation~\eqref{#1}}

\newcommand{\tab}[1]{Table~\ref{#1}}
\newcommand{\alg}[1]{Algorithm~\ref{#1}}
\newcommand{\algline}[1]{Line~\ref{#1}}

\icmltitlerunning{Fast Neural Network Verification via Shadow Prices}

\begin{document}

\twocolumn[
\icmltitle{Fast Neural Network Verification via Shadow Prices}

% It is OKAY to include author information, even for blind
% submissions: the style file will automatically remove it for you
% unless you've provided the [accepted] option to the icml2018
% package.

% List of affiliations: The first argument should be a (short)
% identifier you will use later to specify author affiliations
% Academic affiliations should list Department, University, City, Region, Country
% Industry affiliations should list Company, City, Region, Country

% You can specify symbols, otherwise they are numbered in order.
% Ideally, you should not use this facility. Affiliations will be numbered
% in order of appearance and this is the preferred way.
\icmlsetsymbol{equal}{*}

\begin{icmlauthorlist}
\icmlauthor{Vicen\c{c} Rubies-Royo}{uc}
\icmlauthor{Roberto Calandra}{fbook}
\icmlauthor{Dusan M. Stipanovic}{uiuc}
\icmlauthor{Claire Tomlin}{uc}
\end{icmlauthorlist}

\icmlaffiliation{uc}{Electrical Engineering and Computer Sciences Dept., University of California at Berkeley, Berkeley, California, USA}
\icmlaffiliation{fbook}{Facebook AI Research,
Menlo Park, California, USA}
\icmlaffiliation{uiuc}{Industrial and Enterprise Systems Engineering Dept., University of Illinois Urbana-Champaign, Champaign, Illinois, USA}

\icmlcorrespondingauthor{Vicen\c{c} Rubies-Royo}{vrubies@eecs.berkeley.edu}

% You may provide any keywords that you
% find helpful for describing your paper; these are used to populate
% the "keywords" metadata in the PDF but will not be shown in the document
\icmlkeywords{Machine Learning, ICML}

\vskip 0.3in
]

% this must go after the closing bracket ] following \twocolumn[ ...

% This command actually creates the footnote in the first column
% listing the affiliations and the copyright notice.
% The command takes one argument, which is text to display at the start of the footnote.
% The \icmlEqualContribution command is standard text for equal contribution.
% Remove it (just {}) if you do not need this facility.

\printAffiliationsAndNotice{}  % leave blank if no need to mention equal contribution
%\printAffiliationsAndNotice{\icmlEqualContribution} % otherwise use the standard text.

\begin{abstract}
%Being able to guarantee stable outputs for deep neural networks is crucial in many mission-critical applications (e.g., self-driving cars).
To use neural networks in safety-critical settings it is paramount to provide assurances on their runtime operation. 
Recent work on ReLU networks has sought to verify whether inputs belonging to a bounded box can ever yield some undesirable output.
Input-splitting procedures, a particular type of verification mechanism, do so by recursively partitioning the input set into smaller sets.
The efficiency of these methods is largely determined by the number of splits the box must undergo before the property can be verified.
In this work, we propose a new technique based on shadow prices that fully exploits the information of the problem yielding a more efficient generation of splits than the state-of-the-art.
Results on the Airborne Collision Avoidance System (ACAS) benchmark verification tasks show a considerable reduction in the partitions generated which substantially reduces computation times. These results open the door to improved verification methods for a wide variety of machine learning applications including vision and control. 
\end{abstract}

\section{Introduction}
\label{sec:introduction}

With the increased deployment of deep neural networks (DNN) in many domains, it is becoming more and more pressing to validate said models. While deep learning has shown great promise in applications such as vision \cite{introlecun2015deep}, reinforcement learning \cite{introsilver2017mastering,intromnih2013playing} or speech recognition \cite{intrograves2013speech,introkim2014convolutional}, there are still many safety-critical applications, such as self-driving \cite{introbojarski2016end}, that will not benefit from these advances until models can be effectively validated. 

% Issue
A limitation of current verification approaches is that they can be very slow. It has been shown that the verification problem for feedforward ReLU networks is NP-Complete \cite{KatzBDJK17}. Therefore, it is paramount to find verification methodologies that can improve upon previous results.

% Contribution
In this paper, we propose an approach to reduce the computational cost of verifying deep ReLU networks without any loss of generality or approximation. 
The main contribution is the use of a more efficient input-splitting algorithm -- using so-called shadow prices -- which allows to more effectively decide how to generate the splits of the input box.
As a result, this algorithm reduces the number of splits used for verification tasks, and thus the memory footprint and computational cost of the overall procedure.
Experimental results on the Airborne Collision Avoidance System~(ACAS) standard benchmark demonstrate that our approach significantly reduces the number of splits generated and the time needed to verify properties.

As the number of machine learning applications grows, verification will become more and more important to guarantee safe behaviors from the learned models. 
Our approach is a small step towards tackling the real-world challenges of efficiently and reliably verifying deep neural networks.

\section{Verification of Neural Networks}
\label{sec:verification_of_relu}

We now formalize the verification problem and discuss related work in this area. 
We then briefly introduce ReLU networks and a few of their properties.

\subsection{Verification Problem}
\label{subsec:verification_problem}

Given a set of possible inputs and a neural network, the problem of verifying whether some input will result in an undesirable output is mathematically analogous to checking whether a set mapped through a function intersects some other set. This input set could, for instance, represent a range of valid operational configurations for the system, whereas the output set could represent dangerous/undesirable conditions.
In reinforcement learning and control, for example, the DNN could represent a controller, the input set would be some set of valid states of the system and the output set could correspond to control actions which are known \textit{a priori} to be unsafe. 
Hence, given a DNN $f(x)$, input set $\mathcal{B}$ and an output set $\mathcal{S}$, the verification problem seeks to answer whether the proposition 
\begin{align}
\label{eq:verification_problem_def}
\forall x \in \mathcal{B}, f(x) \notin \mathcal{S}
\end{align}
is true or false.

% ================================

\subsection{Prior Work}
\label{subsec:prior_work}

Starting with the works from \cite{HuangKWW16,KatzBDJK17}, the authors use Satisfiability Modulo Theory~(SMT) solvers to answer \eq{eq:verification_problem_def} for ReLU networks and, more generally, networks containing piece-wise linear activations. 
In their approaches, an answer is reached by leveraging the finite set of possible activations induced by the network's non-linearities. 
A similar reasoning is found in \cite{LomuscioM17,Tjeng} using Mixed Integer Linear Programming~(MILP) solvers. 
Other approaches inspired by reachability include the work from \cite{WeimingXiangReLU}, where the structure of the domain induced by the ReLU non-linearities is exploited to compute the exact set of possible outputs. 
In contrast, in \cite{WeimingXiang}, the set of possible outputs is approximated by gridding the input set. 

Of particular interest for this paper are the works of \cite{Ehlers17} and \cite{WongKolter18}. 
\citet{Ehlers17} introduced a novel convex relaxation of the ReLU non-linearity in order to render the problem easier for the SMT solver. 
\citet{WongKolter18} used this relaxation and duality to compute rapid over-approximations of the set of possible outputs for the network. In \cite{AditiCertified} they expand on this duality approach.
A new interesting direction which does not rely on SMT or MILP solvers was presented by \citet{Shiqi1}. 
In this work, a divide and conquer approach was used to repeatedly partition the input set into smaller sub-domains and check the property individually for each partition.

In our work, we build upon the input-splitting technique from \cite{Shiqi1}. 
We show experimentally that splits based on input-output gradient metrics are in general inefficient. We provide a new methodology that substantially reduces the number of splits and the runtime required to verify a network for a given input set and property.

\subsection{ReLU Network: A Piece-wise Affine Function}
\label{subsec:relu_definition}

Let $f_\theta(x)$ be a ReLU network defined by
\begin{equation}
\label{eq:relu}
\begin{split}
  & \hat{z}_{i+1} = W_i z_i + b_i\,, ~ \text{for} ~ i=1,...,K-1 \\
  & z_j = \max\{\hat{z}_j,0\}\,,  ~ \text{for} ~ j=2,...,K-1 \\
\end{split}
\end{equation}
with $W_i \in \mathbb{R}^{n_{i+1} \times n_i}$, $b \in \mathbb{R}^{n_{i+1}}$, $z_1 = x \in \mathcal{B}$, where $\mathcal{B}$ is a bounded set in $\mathbb{R}^{n_1}$ which we assume to be a box, and $f_\theta(x) = \hat{z}_{K}$. 
The set $\theta = \{W_i,b_i\}_{i=1,...,K-1}$ represents the set of parameters of the network.

\begin{figure}[t]
\centering
\includegraphics[width=\columnwidth]{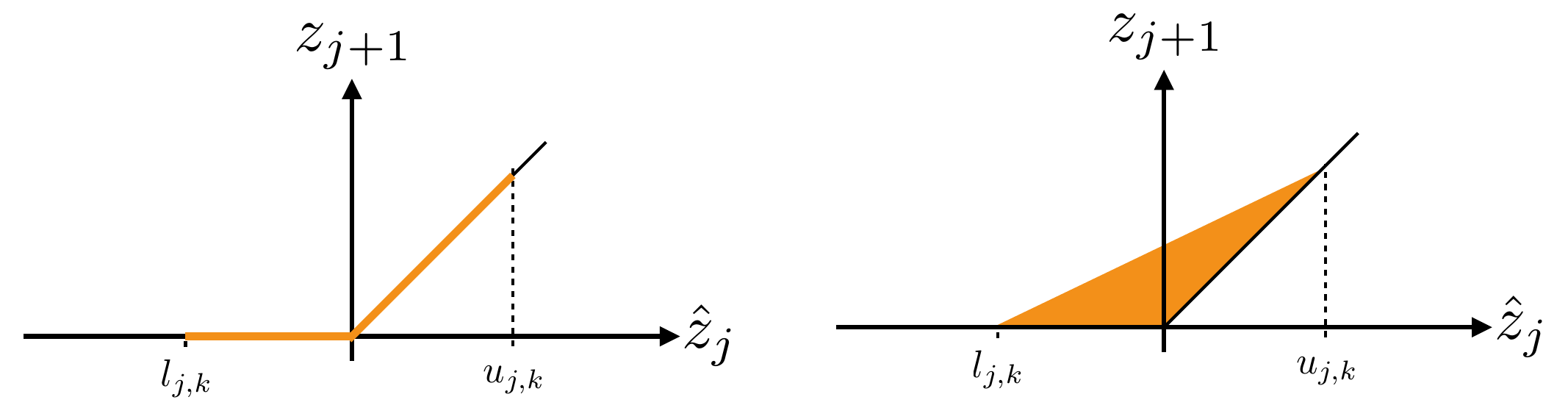} %width=\columnwidth,
\caption{\textit{(Left)} Bounded output of a ReLU node. \textit{(Right)} Convex envelope.}
%\label{fig:convexenvelope}
%\vspace{-10pt}
\label{fig:relu_envelope}
\end{figure}

From the definition of \eq{eq:relu}, it is clear that ReLU networks are piece-wise affine functions. 

This follows from the fact that any composition of an affine function (i.e. linear function plus a bias term) and a piece-wise affine function results in a piece-wise affine function. 
An important aspect that will come into play in later sections, is that the domain is partitioned into polytopic regions $\mathcal{P}_i$ such that $\bigcup_{i} \mathcal{P}_i = \mathbb{R}^{n_1}$. For a closer look at the properties of ReLU networks, including the partition of the domain into polytopic regions, we direct the reader to \cite{linmontufar2014number,linthiago,linraghu2017expressive}.

\section{Overview of Verification via Input-Splits}
\label{sec:verif_recursive_cuts}

We now outline how the verification task is accomplished by iteratively splitting sections of the input set. 
First, we introduce the concept of convex relaxations of ReLU networks to over-approximate the image of the input set as introduced in \cite{Ehlers17} and in \cite{WongKolter18}. 
Next, we show how these relaxations can be used to verify properties of the image, along the lines of the work by \citet{Shiqi1}.

\subsection{Convex Over-approximation of the Image}
\label{subsec:conv_over}

\begin{figure}[t]
\centering
\includegraphics[width=.8\columnwidth]{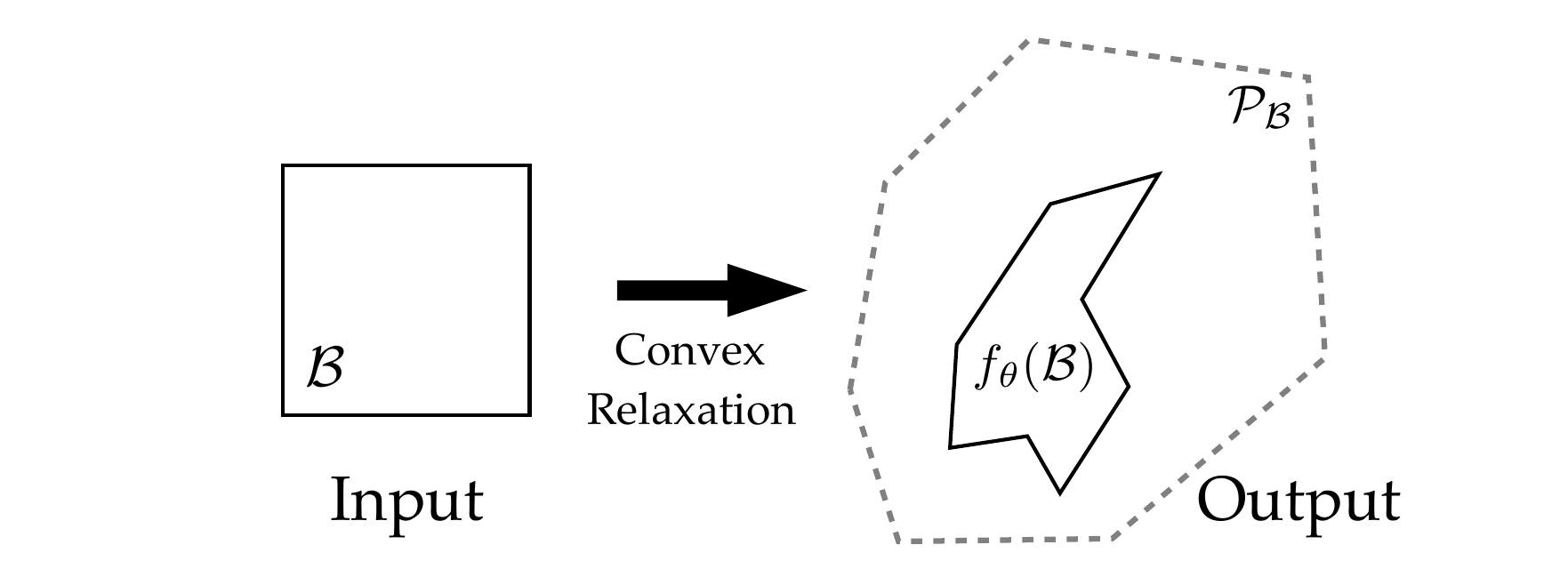} %width=\columnwidth,
\caption{Example of convex over-approximation of the output set.}
\label{fig:convex_overapprox}
\vspace{-10pt}
%\label{fig:relu_overapprox}
\end{figure}

To obtain an over-approximation of the image $f_{\theta}(\mathcal{B})$, it suffices to replace the ReLU non-linearity $\max\{z,0\}$ in each layer by its convex envelope,
\begin{equation}
\label{eq:relaxed-relu}
\begin{split}
 &\hat{z}_{i+1} = W_i z_i + b_i\,, ~ \text{for}~i=1,...,K-1\\
 &z_j \geq 0\,, \quad \text{for}~j=2,...,K-1\\
 &z_j \geq \hat{z}_{j}\,,\\
 &z_j \leq D_j\hat{z}_{j} + d_{j}\,.\\
\end{split}
\end{equation}

The matrix $D_j$ is diagonal, so that ${D_j = diag(c_j)}$, where $c_j$ is a vector of the form ${c_{j,k} = \frac{u_{j,k}}{u_{j,k} - l_{j,k}}}$ and ${{d_j}_k = -\frac{u_{j,k}l_{j,k}}{u_{j,k} - l_{j,k}}}$. The terms $l_{j,k}$ and $u_{j,k}$, which we will shortly define, denote the lower and upper bounds for the $k$-th activation in layer $j$. For \eqref{eq:relaxed-relu}, ${\hat{z}_{K} \in \mathcal{P}_{\mathcal{B}}}$, where $\mathcal{P}_{\mathcal{B}}$ is a bounded convex polytope satisfying ${f_\theta(\mathcal{B}) \subseteq \mathcal{P}_{\mathcal{B}}}$. 
\fig{fig:relu_envelope} shows a typical convex envelope. 
\fig{fig:convex_overapprox} shows how the convex relaxation results in a polytopic over-approximation of the image.

Computing the lower and upper bounds $l_{j,k}$ and $u_{j,k}$, can be accomplished in a layer-by-layer fashion by solving linear programs (LPs) from $j=1$ to $K-2$ of the form 

\begin{equation}
\label{eq:LPs}
\begin{split}
 & l_{j,k}=\min_{\underline{z}} ~(w_{j,k}\underline{z} + b_{j,k}) \quad \text{s.t.} ~ \tilde{A}_j\underline{z} \preceq \tilde{b}_j\,,\\
 & u_{j,k}=\max_{\bar{z}} ~(w_{j,k}\bar{z} + b_{j,k}) \quad \text{s.t.} ~ \tilde{A}_j\bar{z} \preceq \tilde{b}_j\,,\\
\end{split}
\end{equation}
where $\tilde{A}_j$ and $\tilde{b}_j$ represent a polytope in a $d$-dimensional space, where $d = \sum_{l=1}^j n_l$. This polytope grows in dimension as a result of taking into account upper and lower bounds of earlier layers. We also implicitly assume for the remainder of the paper that positive lower bounds or negative upper bounds are automatically set to $0$.  The terms $w_{j,k}$ and $b_{j,k}$ denote the $k$-th row/entry of $W_j$ and $b_j$. In \sec{sec:analytic_gradients}, we study the structure of the constraint set in \eq{eq:LPs} in depth.

\subsection{Image Verification through Refinements}
\label{subsec:im_refinement}

One of the useful features of this bounded convex over-approximation is that it provides sufficient conditions to check whether 
the property of interest can ever be satisfied 
(i.e., from Figure \ref{fig:convex_overapprox}, if the property does not hold for the over-approximation, it can't hold for the image either). However, when properties do hold for the over-approximation nothing can be established with regards to the image.

\subsubsection{Recursively Splitting Sets}
\label{subsubsec:recursive_splitting}

In \sec{subsec:conv_over}, we have explained how to build a convex over-approximation of the image in a layer-by-layer basis. 
Note, that for any split of the input set into two subsets $\mathcal{B}_1$ and $\mathcal{B}_2$ such that $\mathcal{B}_1 \cup \mathcal{B}_2 = \mathcal{B}$, it holds that
\begin{equation}
 \mathcal{P}_{\mathcal{B}_1} \cup \mathcal{P}_{\mathcal{B}_2} \subseteq \mathcal{P}_{\mathcal{B}}\,.
\end{equation}
This follows from the fact that splitting the input set reduces all the feasible regions for the LPs in \eq{eq:LPs}, which results in greater lower bounds or smaller upper bounds in the computation of $\mathcal{P}_{\mathcal{B}_1}$ and $\mathcal{P}_{\mathcal{B}_2}$.

Splitting the input set is particularly useful for verification since it breaks the problem into two sub-problems. Specifically, and without loss of generality, one of three things may happen:

\begin{enumerate}[topsep=0pt,itemsep=0pt,partopsep=1ex,parsep=1ex]
    \item The property does not hold for either $\mathcal{P}_{\mathcal{B}_1}$ nor $\mathcal{P}_{\mathcal{B}_2}$, and thus the property is not satisfied by the image.
    \item The property of interest holds for $\mathcal{P}_{\mathcal{B}_1}$ but does not hold for $\mathcal{P}_{\mathcal{B}_2}$, in which case $\mathcal{B}_2$ can be discarded from the verification problem.
    \item Finally, both $\mathcal{P}_{\mathcal{B}_1}$ and $\mathcal{P}_{\mathcal{B}_2}$ satisfy the property and nothing can be established.
\end{enumerate}

Given these outcomes, a natural algorithm arises for the verification of the property of interest; starting with $\mathcal{B}$, we compute the convex over-approximation $\mathcal{P}_{\mathcal{B}}$. If the property does not hold for $\mathcal{P}_{\mathcal{B}}$, the property does not hold for the image and we are done. Otherwise, we can split $\mathcal{B}$ in two halves $\mathcal{B}_1$ and $\mathcal{B}_2$, and compute $\mathcal{P}_{\mathcal{B}_1}$ and $\mathcal{P}_{\mathcal{B}_2}$. 
Given the list of possible outcomes, the algorithm will either: end with the property being false, be able to discard one of the halves, or, in the worst case, keep both $\mathcal{B}_1$ and $\mathcal{B}_2$ for further analysis. This procedure induces a growing \textit{binary tree} whose nodes represent smaller and smaller regions of the input set $\mathcal{B}$.

\begin{algorithm}[t]
\caption{Recursive Splitting (Depth First Search)}\label{alg:abstract_algo}
\begin{algorithmic}[1]
\Procedure{Verification}{$\mathcal{S},\mathcal{B},\theta$}
\State $\mathcal{P}_{\mathcal{B}} \gets \text{ConvexOverApprox}(\mathcal{B},\theta)$
\If {$\mathcal{P}_{\mathcal{B}} \cap \mathcal{S} = \emptyset$} \label{l_alg:ifcondcap}
\State \Return False
\Else
\If {$\text{IsExact}(\mathcal{P}_{\mathcal{B}})$} \label{l_alg:ifexact} 
\State \Return True
\EndIf
\State $\mathcal{B}_1,\mathcal{B}_2 \gets \text{Split}(\mathcal{B})$ \label{l_alg:split}
\State \Return $\text{Verification}(\mathcal{S},\mathcal{B}_1,\theta) \lor$ 
\State \quad \quad \quad $\text{Verification}(\mathcal{S},\mathcal{B}_2,\theta)$
\EndIf
\EndProcedure
\end{algorithmic}
\end{algorithm}

\alg{alg:abstract_algo} provides the aforementioned procedure for verifying whether the image of $\mathcal{B}$ intersects with a given set $\mathcal{S}$, which we assume to be a union of a finite number convex sets. 
This assumption is required so that in \algline{l_alg:ifcondcap} the intersection check can be done efficiently via convex optimization. When the intersection is non-empty, in \algline{l_alg:ifexact}, we check whether the over-approximation is tight/exact or not. If it is, it must be true that the input set satisfies the property. 
If it is not, in line \ref{l_alg:split} we split the set into two halves and attempt to solve the verification problem for each half separately. Sections \ref{subsubsec:image_equivalence} and \ref{subsubsec:splitting_criterion} describe the auxiliary functions ``IsExact" and ``Split" in lines \ref{l_alg:ifexact} and \ref{l_alg:split}.

\subsubsection{Over-approximation and Image Equivalence}
\label{subsubsec:image_equivalence}

In the previous section we mentioned that some instances of $\mathcal{P}_\mathcal{B}$ can be checked for tightness or exactness, that is $\mathcal{P}_\mathcal{B} \equiv f_\theta(\mathcal{B})$. 
From \sec{subsec:relu_definition}, we know that a ReLU network sub-divides the domain into convex polytopes $\mathcal{P}_i$, and that within each polytope the input-output relation is affine, i.e. it is a piece-wise affine function. Then the following must hold:

\begin{equation}
\label{eq:equivalence}
\begin{split}
    & 1.) ~\mathcal{B} \subseteq \mathcal{P}_i \iff u_{j,k} \leq 0 ~~\text{or}~~ 0 \leq l_{j,k} ~ \forall j,k\\
    & 2.) ~\mathcal{B} \subseteq \mathcal{P}_i \implies \mathcal{P}_\mathcal{B} \equiv f_\theta(\mathcal{B})
\end{split}
\end{equation}

The first line in \eq{eq:equivalence} follows from the fact that the boundaries of the polytopes $\mathcal{P}_i$ arise from the discontinuity of the non-linear activations $\max\{z,0\}$. The second line follows from the first one: if $l_{j,k} \leq u_{j,k} \leq 0$ or $0 \leq l_{j,k} \leq u_{j,k}$ for all $j$ and $k$, then the relaxations of the ReLU activations will be exact and $f_\theta$ is just an affine map.

\subsubsection{Splitting Criterion}
\label{subsubsec:splitting_criterion}

\begin{figure*}[t]
%\begin{center}
\centerline{\includegraphics[width=0.8\linewidth]{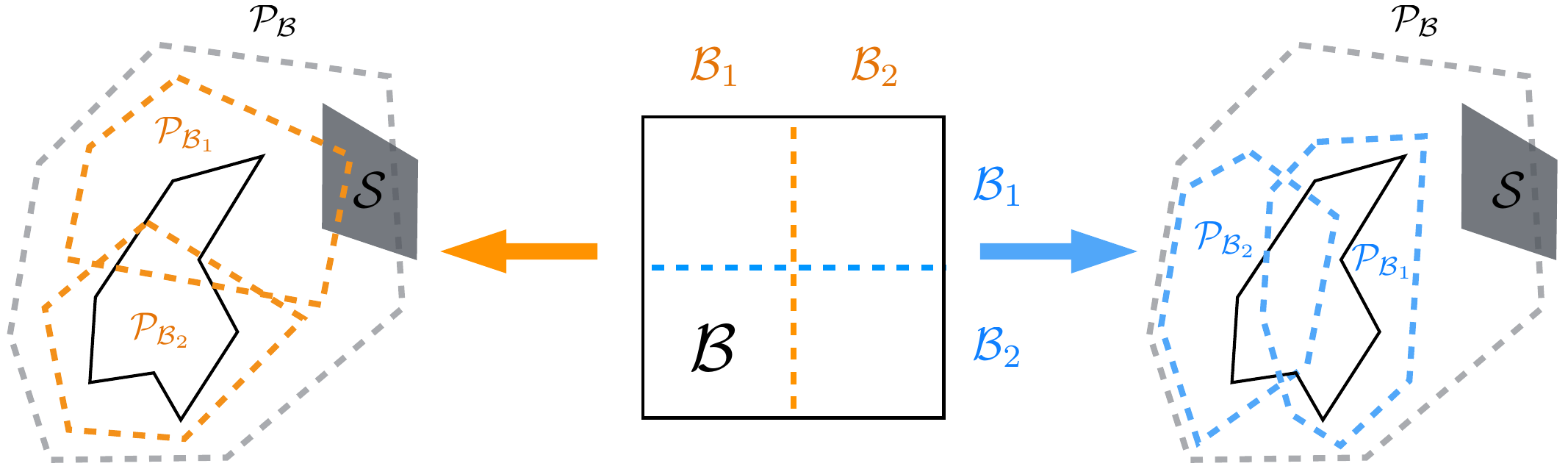}} %width=\columnwidth,
\caption{The choice of axis along which the set $\mathcal{B}$ is split may affect the verification time. In this example, a vertical split (orange) results in two over-approximations, one of which still intersects with $\mathcal{S}$, whereas the horizontal split (blue) results in tighter over-approximations.}
\label{fig:split_consequences}
%\end{center}
\end{figure*}

From \alg{alg:abstract_algo}, a natural question arises: what is considered to be a ``good" split of the input set? While easy to state, this question is far from trivial. 
Picking appropriate splits has important consequences regarding the time and memory efficiency of input-splitting verification algorithms. 
In \fig{fig:split_consequences}, an example is provided showcasing this phenomenon. The horizontal split results in an over-approximation that guarantees that the image of $\mathcal{B}$ does not intersect with $\mathcal{S}$. 
In contrast, the vertical split results in an over-approximation that still intersects $\mathcal{S}$.

To the best of our knowledge, current input-splitting methods \cite{Shiqi1} use gradient information between inputs and outputs to decide which axis to split. 
These methods leverage the structure of the Jacobian of the ReLU network to compute bounds on the gradient between inputs and outputs. 
In particular, note that for any point in the domain, the Jacobian for a ReLU network can always be written as
\begin{equation}
J = W_{K-1}\prod_{i=1}^{K-2} \Sigma_iW_{i}
\end{equation}
for $\Sigma_i = diag(v_i)$ and the Boolean vector $v_i \in \{0,1\}^{n_{i+1}}$. Starting from $i=K-1$ going backwards to $i=1$, these methods select appropriate values of $v_i$ to compute upper bounds for $||\frac{df_\theta(x)}{d x_{k}}||_{\infty} \leq U_k$ for $k=1,...,n_1$. Using the side lengths $\Delta x_k$ of the box, $\mathcal{B}$ is split in half across the axis with greatest smear value $s_k = U_k\Delta x_k$. Geometrically, this mechanism tries to reduce in half the box along the axis that causes most ``stretching'' for any of the outputs. This reasoning, however, can be counterproductive when considering which splits to make. In particular, the upper bound might be very loose in practice, and even in cases where $U_k$ can be achieved, it may be the case that the polytopic region that achieves this upper bound is very small. 

Fundamentally, the over-approximation is caused by the convex relaxations of the ReLU non-linearities, therefore, we posit that an effective splitting procedure should leverage the information of the relaxed nodes (i.e. nodes for which the convex relaxation is not exact) rather than using bounds on the gradients between input and outputs. In the next sections we investigate how to estimate changes in the upper and lower bounds of any given node given a specific split in $\mathcal{B}$.

\section{Shadow Prices and Bound Rates}
\label{sec:analytic_gradients}

We now study the sensitivity of the lower and upper bounds for a relaxed ReLU node with respect to changes in the shape of the input box. 
To that end, we introduce some useful properties of linear programs and relate them to our problem at hand.

\subsection{Measuring Constraint Sensitivity}
\label{subsec:duals_and_sens}

For a linear program with non-empty feasible region the following property holds
\begin{align}
\label{eq:Lagrange_LPs}
  &\underline{p}^* = \min_{s.t. ~ A\underline{z} \preceq b} ~ w\underline{z} \quad = \quad ~ w\underline{z}^* + (A\underline{z}^* - b)^T\underline{\lambda}^*\,\\
 &\bar{p}^* = \max_{s.t. ~A\bar{z} \preceq b} ~ w\bar{z} \quad = \quad ~ w\bar{z}^* + (b - A\bar{z}^*)^T\bar{\lambda}^*\,,
\end{align}
where $ \bar{z}^{*} $ and $ \underline{z}^{*} $ correspond to the primal's maximizer/minimizer and $ \bar{\lambda}^{*} \succeq 0 $ and $ \underbar{$\lambda$}^{*} \succeq 0 $ correspond to the dual's minimizer/maximizer. 
This result follows from strong duality and complementary slackness \cite{boyd2004convex}. 
A useful feature of the right-hand sides of \eq{eq:Lagrange_LPs} is that they link how small perturbations of the constraint parameters ($A$ and $b$) affect the optimal values $\bar{p}^*$ and $\underline{p}^*$. 
In the field of economics, the rate of increase/decrease of the optimal value with respect to a certain constraint is known as the \textit{shadow price}. In some instances the shadow prices are the optimal dual variables.   

From \eqref{eq:Lagrange_LPs}, we can readily study how small changes in the size of the input box affect the upper and lower bounds of all nodes in the first layer of the network. In particular, it follows that for the $k$-th node in the first layer and the $i$-th bias of our constraint set,
\begin{equation}
\label{eq:first_layer_sens}
\frac{d l_{1,k}}{d \tilde{b}_{1,i}} = -\underline{\lambda}_{k,i}^* \quad  \quad \frac{d u_{1,k}}{d \tilde{b}_{1,i}} = \bar{\lambda}_{k,i}^*.
\end{equation}
This result is expected, since the growth of bias terms results in a bigger box and, thus, in smaller lower bounds and bigger upper bounds. A nice feature of most modern LP solvers is that they provide the associated optimal dual variables when solving the primal problem. These rates can therefore be computed without any noticeable computational overhead when generating the convex over-approximations.

To derive the rates of the upper and lower bounds for all nodes in the network we first need to fully characterize the constraint sets given by $\tilde{A}_{1:K-1}$ and $\tilde{b}_{1:K-1}$ first introduced in \sec{subsec:conv_over}.

\subsection{Constraint Sets Characterization}
\label{subsec:constraint_char}

Thus far we have seen how the rates for the lower and upper bounds can be computed for the first layer. For this particular instance, the constraint set is given by $\tilde{A}_1$ and $\tilde{b}_1$, and together they represent a box in $\mathbb{R}^{n_1}$ which is provided to us. The expressions for $\tilde{A}_j$ and $\tilde{b}_j$ for the $j$-th ($j \geq 2$) layer are given by
\begin{equation}
\label{eq:A_b}
\begin{split}
& \tilde{A}_j = 
\begin{bmatrix}
\tilde{A}_{j-1} & 0\\
 O_{j-2} & R_{j-1}\\
\end{bmatrix}, 
R_{j-1} = 
\begin{bmatrix}
0 & -I\\
W_{j-1} & -I\\
-D_{j-1}W_{j-1} & \phantom{-}I\\
\end{bmatrix},\\
& \tilde{b}_j = 
\begin{bmatrix}
\tilde{b}_{j-1} \\ r_{j-1}
\end{bmatrix}, 
r_{j-1}^T = 
%\begin{bmatrix}
 [0^T  -b_{j-1}^T ~~ (D_{j-1}b_{j-1} + d_{j-1})^T]
%\end{bmatrix},
\end{split}
\end{equation}
where $O_{j-2}$ is a matrix of zeros whose number of columns is equal to $\sum_{k=1}^{j-2} n_k$. 
Note how the dimensionality of the constraints given by $\tilde{A}_j$ and $\tilde{b}_j$ increases as we try to compute upper and lower bounds for deeper layers. 
These expressions can be derived from the inequalities provided in \eq{eq:relaxed-relu}. As examples, we provide the expressions for $\tilde{A}_{2:3}$ and $\tilde{b}_{2:3}$ 

\begin{equation}
\begin{split}
\tilde{A}_2 &=
\begin{bmatrix}
\tilde{A}_1 & 0\\
0 & -I\\
W_{1} & -I\\
-D_{1}W_{1} & \phantom{-}I\\
\end{bmatrix},
\tilde{b}_2 = 
\begin{bmatrix}
\tilde{b}_{1} \\ 0 \\ -b_1 \\ D_{1}b_{1} + d_{1}
\end{bmatrix}, 
\\
\tilde{A}_3 &=
\begin{bmatrix}
\tilde{A}_1 & 0 & 0\\
0 & -I & 0\\
W_{1} & -I & 0\\
-D_{1}W_{1} & \phantom{-}I & 0\\
0 & 0 & -I\\
0 & W_{2} & -I\\
0 & -D_{2}W_{2} & \phantom{-}I\\
\end{bmatrix},
\tilde{b}_2 = 
\begin{bmatrix}
\tilde{b}_{1} \\ 0 \\ -b_1 \\ D_{1}b_{1} + d_{1} \\ 0 \\ -b_2 \\ D_{2}b_{2} + d_{2}
\end{bmatrix}.
\\
\end{split}
\end{equation}

\subsection{Forward Computation of the Bound Rates}
\label{subsec:constraint_char}

With the definitions for $\tilde{A}_j$ and $\tilde{b}_j$ available and \eqref{eq:Lagrange_LPs}, we can proceed to compute the rates of the upper and lower bounds with respect to the bias terms of our input box. Denoting $(\underline{\lambda}_{j,k}^*,\bar{\lambda}_{j,k}^*)$ and $(\underline{z}_{j,k}^*,\bar{z}_{j,k}^*)$ as the set of primal and dual optimal variables for the maximization and minimization problems from \eqref{eq:LPs}, we have that

\begin{equation}
\label{eq:rates_general}
\begin{split}
\frac{dl_{j,k}}{d\tilde{b}_{1,i}} &= \frac{d}{d\tilde{b}_{1,i}}\left(\tilde{b}_j^T\underline{\lambda}_{j,k}^*\right) - \frac{d}{d\tilde{b}_{1,i}} \left((\tilde{A}_j \underline{z}_{j,k}^*)^T \underline{\lambda}_{j,k}^*\right)\\
\frac{du_{j,k}}{d\tilde{b}_{1,i}} &= - \frac{d}{d\tilde{b}_{1,i}}\left(\tilde{b}_j^T \bar{\lambda}_{j,k}^*\right) + \frac{d}{d\tilde{b}_{1,i}}\left((\tilde{A}_j \bar{z}_{j,k}^*)^T\bar{\lambda}_{j,k}^*\right).
\end{split}
\end{equation}

Before proceeding to drive the analytic expression of \eq{eq:rates_general}, it is worth noting that changes in the bias terms of the input box $\tilde{b}_{1,:}$ only affect (through the computation of the upper and lower bounds) a subset of the entries of $\tilde{A}_j$ and $\tilde{b}_j$. 
In particular, given $R_{1:j-1}$ and $r_{1:j-1}$ defined in \eq{eq:A_b}, only entries containing dependencies with $D_{1:j-1}$ and $d_{1:j-1}$ will contribute to the gradient expression in $\eqref{eq:rates_general}$. 
Hence, focusing our attention to the first term, for any vector $\lambda$
\begin{equation}
\label{eq:rates_first_term}
\begin{split}
\frac{d}{d\tilde{b}_{1,i}}(\tilde{b}_j^T \lambda) & = \frac{d}{d\tilde{b}_{1,i}}\left(\tilde{b}_1^T \lambda_{0} + \sum_{l = 1}^{j-1} r_{l}^T \lambda_l\right)\\
& = \lambda_{0,i} + \sum_{l = 1}^{j-1} \frac{d}{d\tilde{b}_{1,i}} (D_{l}b_{l} + d_{l})^T \lambda_{\hat{l}}\\
& = \lambda_{0,i} + \sum_{l = 1}^{j-1} \sum_{t=0}^{n_l}  \frac{d}{d\tilde{b}_{1,i}} \frac{u_{l,t}b_{l,t} - u_{l,t}l_{l,t}}{u_{l,t} - l_{l,t}} \lambda_{\hat{l},t}\\
& = \lambda_{0,i} + \sum_{l = 1}^{j-1} \sum_{t=0}^{n_l}  \left(c_1\frac{du_{l,t}}{d\tilde{b}_{1,i}} + c_2\frac{dl_{l,t}}{d\tilde{b}_{1,i}}\right)\lambda_{\hat{l},t}
\end{split}
\end{equation}
where $c_1 = \frac{l_{l,t}(l_{l,t} - b_{l,t})}{(u_{l,t} - l_{l,t})^2}$ and $c_2 = \frac{u_{l,t}(b_{l,t} - u_{l,t})}{(u_{l,t} - l_{l,t})^2}$ and $\lambda_{\hat{l}},\lambda_{l}$ correspond to a subset of the entries in $\lambda_l$ and $\lambda$ respectively. The term $b_{l,t}$ denotes entry $t$ of the bias term in layer $l$.
From \eq{eq:rates_first_term} it is clear that the rates of the upper and lower bounds for layer $j$ depend on all the rates from previous layers.

The rates for the second term can be derived in a similar manner. For any vector $z$ and $\lambda$,
\begin{equation}
\label{eq:rates_second_term}
\begin{split}
\frac{d}{d\tilde{b}_{1,i}} (\tilde{A}_j z)^T \lambda & = \frac{d}{d\tilde{b}_{1,i}}((\tilde{A}_1 z_0)^T \lambda_{0} + \sum_{l = 1}^{j-1} (R_{l}\begin{bmatrix}z_{l-1}\\z_{l}\end{bmatrix})^T \lambda_l)\\
& = \sum_{l = 1}^{j-1} \frac{d}{d\tilde{b}_{1,i}} \left(-D_{l}W_{l}z_{l-1}\right)^T \lambda_{\hat{l}}\\
& = -\sum_{l = 1}^{j-1} \sum_{t=0}^{n_l}  \frac{d}{d\tilde{b}_{1,i}} \frac{(w_{l,t}z_{l-1})u_{l,t}}{u_{l,t} - l_{l,t}} \lambda_{\hat{l},t}\\
& = \sum_{l = 1}^{j-1} \sum_{t=0}^{n_l}  \left(\hat{c}_1\frac{dl_{l,t}}{d\tilde{b}_{1,i}} - \hat{c}_2\frac{du_{l,t}}{d\tilde{b}_{1,i}}\right)\lambda_{\hat{l},t}
\end{split}
\end{equation}
where $\hat{c}_1 = \frac{u_{l,t}(w_{l,t}z_{l-1})}{(u_{l,t} - l_{l,t})^2}$ and $\hat{c}_2 = \frac{l_{l,t}(w_{l,t}z_{l-1})}{(u_{l,t} - l_{l,t})^2}$, and $w_{l,t}$ corresponds to row $t$ of the weight matrix in layer $l$. 

Using \eqref{eq:rates_first_term} together with \eqref{eq:rates_second_term} we can compute expressions for \eqref{eq:rates_general} in a forward manner using the optimal primal and dual variables.

\section{Bound Estimation and Splitting}
\label{sec:prediction_based_splitting}

\begin{algorithm}[t]
\caption{Bound Estimation-based Splitting}\label{alg:estimation_algo}
\begin{algorithmic}[1]
\Procedure{Split}{$\mathcal{B},({l},{u}),(\underline{z}^*,\underline{\lambda}^*),(\bar{z}^*,\bar{\lambda}^*)$}
\State $\frac{dl_{j,k}}{d\tilde{b}_{1,i}} \gets \text{LBounds}(\mathcal{B},({l},{u}),(\underline{z}^*,\underline{\lambda}^*),(\bar{z}^*,\bar{\lambda}^*))$
\State $\frac{du_{j,k}}{d\tilde{b}_{1,i}} \gets \text{UBounds}(\mathcal{B},({l},{u}),(\underline{z}^*,\underline{\lambda}^*),(\bar{z}^*,\bar{\lambda}^*))$
\State $i=1,c=L(1)$
\For {$k = 2,...,n_1$}
\If {$L(k) < c$}
\State $c \gets L(k)$
\State $i \gets k$
\EndIf
\EndFor
\State $\mathcal{B}_1 \gets \mathcal{B}_i$
\State $\mathcal{B}_2 \gets \mathcal{B} $\textbackslash$ \mathcal{B}_i$
\State \Return $\mathcal{B}_1,\mathcal{B}_2$
\EndProcedure
\end{algorithmic}
\end{algorithm}

With the information provided in \sec{sec:analytic_gradients}, we have means to estimate how the lower and upper bounds of our convex relaxation change as a function of the biases of the input set. 
In particular, splitting the box $\mathcal{B}$ in half can be viewed as translating one the facets of the box to its center. Since the translation of any facet is achieved by reducing the associated bias term by a certain amount $\Delta \tilde{b}_{1,i}$, the estimated new lower and upper bounds for the resulting set $\mathcal{B}_i \subset \mathcal{B}$ will be
\begin{equation}
\label{eq:estimated_uandl}
\begin{split}
 & l_{j,k}^{\mathcal{B}_i} \approx l_{j,k} + \frac{d l_{j,k}}{db_{1,i}}\Delta \tilde{b}_{1,i}\\
 & u_{j,k}^{\mathcal{B}_i} \approx u_{j,k} + \frac{d u_{j,k}}{db_{1,i}}\Delta \tilde{b}_{1,i}\,.\\
\end{split}
\end{equation}
Using these estimates, we propose the following metric for determining which axis to split along:
\begin{equation}
\label{eq:metric_for_split}
\begin{split}
 & L(i) = -\sum_{j=1}^{K-2} \sum_{k=1}^{n_j} \max\{0,u_{j,k}^{\mathcal{B}_i}\}\min\{0,l_{j,k}^{\mathcal{B}_i}\}.
\end{split}
\end{equation}
Whichever axis minimizes $L(i)$ is chosen for splitting. The max/min terms in the sum ensure that upper and lower bounds that start close to zero have less contribution in the overall splitting decision. Note how the cost metric is only zero whenever all relaxations are tight. 
We summarize this splitting procedure in \alg{alg:estimation_algo}.

\section{Experiments}
\label{sec:experiments}

\begin{figure*}[h!]
\label{fig:barplots}
\centering
\includegraphics[width=\textwidth]{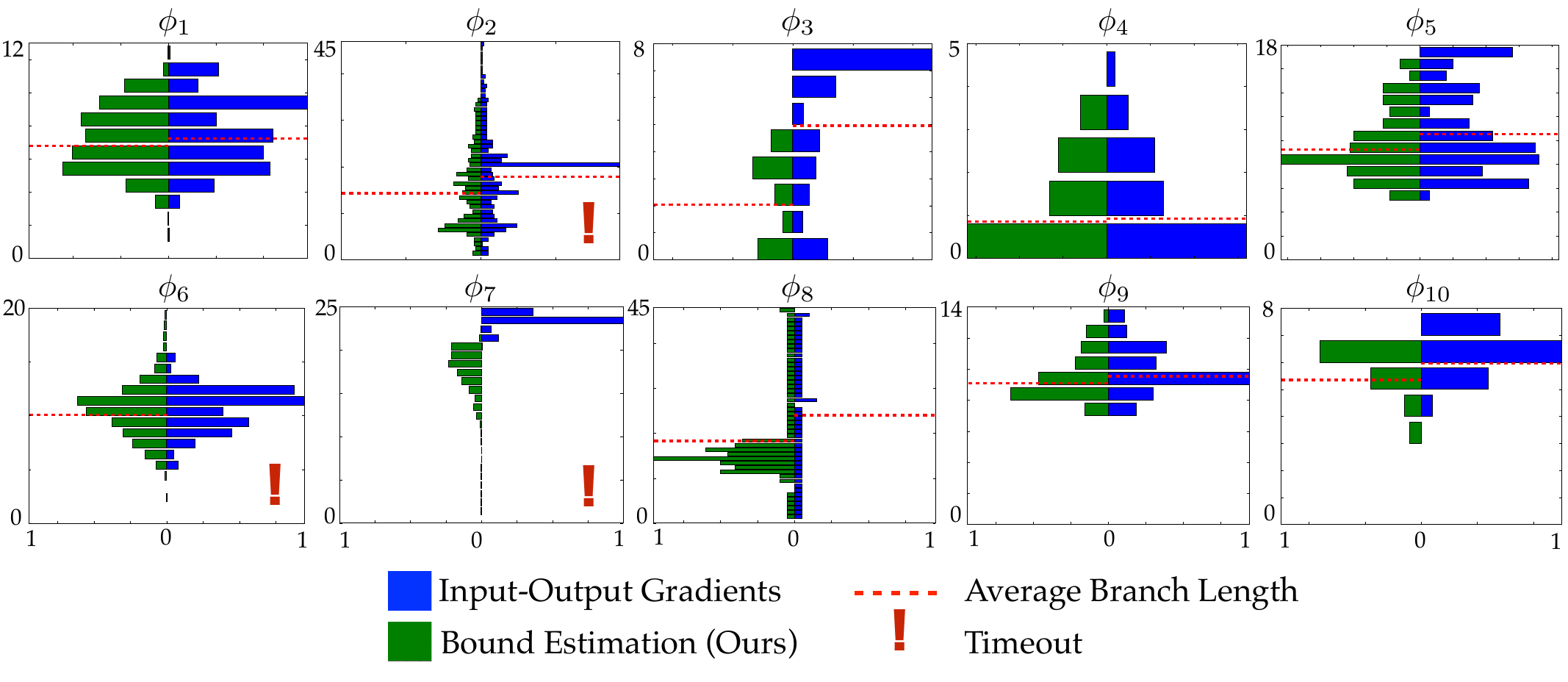}
\caption{Horizontal histograms displaying the number of branches of each length generated by each type of splitting procedure. Each pair of histograms is normalized with the maximum branch length reached for that specific property. For histograms where timeouts occur we do not report the average branch length, since the associated tree has not finished growing. For property 2 we do report the average branch length only for networks that could be verified by both IOG and BE procedures.
}
\label{fig:barplots}
\end{figure*}

In this section, we present the experimental results obtained on a set of benchmark verification tasks by our input-splitting approach based on bound rates.
As baseline, we compare against the input-output gradient-based method discussed in \sec{subsubsec:splitting_criterion}. 

\subsection{Airborne Collision Avoidance System Verification}
\label{subsec:ACAS}

The ACAS benchmark verification task comprises a set of ten properties $\phi_{1:10}$ to be checked on a subset of 45 feedforward ReLU networks. All of the neural networks have the same architecture, with 5 inputs, 5 outputs and 6 hidden layers with 50 neurons each. 
The five inputs represent a specific configuration between two aircraft, one which is denoted as the \textit{ownship}, and the other as the \textit{intruder}. The inputs are $\rho$ (distance between aircraft), $\theta$ (heading angle of ownship), $\psi$ (heading angle of intruder), $v_{own}$ (speed of ownship) and $v_{int}$ (speed of intruder). The output corresponds to five scalars: $COC$ (Clear of Conflict), \textit{weak left}, \textit{weak right}, \textit{strong left} and \textit{strong right}. The output with the greatest value is the advice action for the ownship. 
The properties $\phi_{1:10}$ specify a box-shaped subset $\mathcal{B}$ in the input space and a set $\mathcal{S}$ in the output space. A property is said to be \textit{unsatisfied} for a given network if \eqref{eq:verification_problem_def} is true, otherwise the network \textit{satisfies} the property. 
Details for each property are given in the Appendix. 
For further details on this benchmark we direct the reader to \cite{KatzBDJK17}.

\subsection{Experimental Details}
\label{subsec:exp_details}

There are a few technical aspects that need to be clarified before proceeding to the results. 
In our implementation, we did \emph{not} use any form of parallelism, even though this approach can be readily extended to a parallel implementation similar to \cite{Shiqi1}. 
Each individual network to be verified was given a maximum execution time of 3 hours, at which point the verification function halts and a timeout flag is returned. 
All verification tasks were performed on a 12 core, 64-bit machine with Intel Core i7-5820K CPUs @ 3.30GHz. 
The experimental setting, and our approach, was coded in Python using the Gurobi optimization package.
%Full code for reproducing the experiments will be published upon acceptance.

\subsection{Comparison of Splitting Procedures}
\label{subsec:comparison_of_splitting}

\begin{table}[t]
%\begin{threeparttable}
\centering
\resizebox{\linewidth}{!}{% 
\begin{tabular}{c| c| c c| c c} 
 \hline
 \multirow{2}{*}{$\phi$} & \multirow{2}{*}{\# NNs}& \multicolumn{2}{c|}{IOG} & \multicolumn{2}{c}{BE (Ours)} \\
  & & U/S/T &   Search depth  & U/S/T & Search depth  \\  \hline \hline
 1 & 45 & 45/0/0 & $7.22 \pm 2.11$ & 45/0/0 & $6.79 \pm 1.90$ \\ 
 2 & 34 & 0/30/4 & $17.87 \pm 9.00^*$ & 0/32/2 & $14.29 \pm 7.94^*$ \\
 3 & 42 & 42/0/0 & $4.96 \pm 2.50$ & 42/0/0 & $2.03\pm1.49$ \\ 
 4 & 42 & 42/0/0 & $0.92\pm1.13$ & 42/0/0 &  $0.85\pm1.03$ \\
 5 & 1 & 1/0/0 & $10.5\pm3.65$ & 1/0/0 & $9.25\pm2.79$ \\ 
 6 & 1 & 0/0/1 & N/A & 1/0/0 & $10.1\pm2.48$ \\ 
 7 & 1 & 0/0/1 & N/A & 0/0/1 & N/A \\ 
 8 & 1 & 0/1/0 & $24.1\pm13.40$ & 0/1/0 & $18.33\pm10.05$ \\ 
 9 & 1 & 1/0/0 & $9.53\pm1.47$ & 1/0/0 & $9.09\pm1.47$ \\
 10 & 1 & 1/0/0 & $5.96\pm0.80$ & 1/0/0 & $5.34\pm0.89$ \\ \hline
\end{tabular}
% \captionof{figure}{your caption text}
%\end{threeparttable}
}
\caption{Verification results (U: unsatisfied. S: satisfied. T: timeout) and search depth (mean $\pm$ standard deviation) for all properties of the ACAS benchmark. These results show that our approach validates the properties as well as, or better than, IOG -- even with a reduced search depth. We use $^*$ in property 2 to denote that the comparison is \emph{only} for networks where neither IOG nor BE procedures had a timeout.}
\label{tbl:averages_stds}
\end{table}

In this section, we provide a side-by-side comparison between input-output gradient-based splits~(IOG) and splits generated by our approach based on bound estimation~(BE). 

\tab{tbl:averages_stds} shows the verification results for each of the 10 properties. The first column enumerates the properties and the second column shows the number of networks the property was tested against. The third and fifth columns show the verification results, which can either be unsatisfied, satisfied or timeout, for the IOG and BE-based splits, respectively. 
Columns four and six show the average depth for the binary trees generated during the verification procedures.
When timeouts occurred, we did not report the average depth and standard deviations, as they are misleading metrics representing partially built binary trees, the only exception being property 2\footnote{Similar to \cite{Shiqi1}, we omit networks $\{N_{4,2},N_{5,3}\}.$}, for which 30 networks were able to be verified by both IOG and BE procedures\footnote{IOG timeouts: networks $\{N_{3,3},N_{3,4},N_{3,8},N_{4,9}\}$. BE timeouts: networks $\{N_{3,3},N_{4,9}\}$.}. In the Appendix we include an additional graph with timeouts for property 2.  

\begin{table}[t]
\centering
\begin{tabular}{c| c |c c| c} 
 \hline
 \multirow{2}{*}{$\phi$} & \multirow{2}{*}{\# NNs} & \multicolumn{2}{c|}{\# Nodes} & \multirow{2}{*}{$t_\text{BE}/t_\text{IOG}$} \\
 %\cline{3-4}
  & & IOG & BE &  \\
 \hline\hline
 1 & 45 & 5241 & 4541 & \textbf{0.873} \\  
 2 & 34$^*$ & 799$^\dagger$ & 553$^\dagger$ & \textbf{0.872}$^\dagger$ \\ 
 3 & 43 & 464 & 164 & \textbf{0.361} \\ 
 4 & 43 & 84 & 82 & \textbf{0.922} \\ 
 5 & 1 & 491 & 369 & \textbf{0.877} \\ 
 6 & 1$^*$ & $>$1808* & 1210 & $<$\textbf{0.682}* \\  
 7 & 1$^*$ & $>$1269* & $>$983* & 1.0* \\  
 8 & 1 & 95 & 245 & 1.215 \\ 
 9 & 1 & 947 & 737 & \textbf{0.791} \\ 
 10 & 1 & 105 & 63 & \textbf{0.771} \\ \hline
\end{tabular}
% \captionof{figure}{your caption text}
\caption{Number of nodes for the two split mechanisms (smaller is better) and the ratio of computational time for the two methods. We can see that for 8 out of 10 properties our approach (BE) reduces the number of nodes generated, as well as the computational time. $^*$ indicates that some verification did not complete within the allocated time. We use $^\dagger$ in property 2 to denote that the comparison is \emph{only} for networks where neither IOG nor BE procedures had a timeout.}
\label{tbl:comparison}
%\vspace{-20pt}
\end{table}

The horizontal histograms in \fig{fig:barplots} depict the number of branches (horizontal axis) of a specific length (vertical axis) for the binary trees generated by the verification tasks. 
For each histogram, the distribution on the left (green) corresponds to the distribution generated by BE-based splitting. 
The distributions on the right (blue) correspond to IOG-based splits. 
The dashed red red line in each plot shows the average depth, which is reported in \tab{tbl:averages_stds}. 
In cases where timeouts occurred for either type of procedure we include an exclamation mark.

\tab{tbl:comparison} shows a few more important metrics from the verification tasks. The first to columns are the same as \tab{tbl:averages_stds}. Columns three and four show the exact number of nodes that were generated during verification of each task. The fifth column shows the ratio of times between BE-based splits and IOG-based splits for a given verification task. Values below 1.0 imply a strict improvement of BE-splits over IOG-splits. In cases where timeouts occurred we include an asterisk symbol.

\section{Discussion}
\label{sec:Disc}

As seen in the previous section in Table \ref{tbl:averages_stds}, BE-based splits are able to prove all properties to be either unsatisfied or satisfied except for $\phi_7$ and for two of the networks in $\phi_2$. These results match the ones found in \cite{KatzBDJK17}, with the exception of $\phi_1$, where we do not timeout\footnote{Reluplex uses 4h timeout threshold. For property 1, Reluplex reported 4 timeouts.}, and $\phi_2$, where we do timeout twice. In contrast, while IOG-based splits are also able to verify most of the properties, additional timeouts occur for properties $\phi_2$ and $\phi_6$. In all cases, we found the average search depth $\mu_{BE}$ to be smaller than $\mu_{IOG}$. Along these same lines, \tab{tbl:comparison} shows that for 8 out of the 10 properties we get a strict improvement on the amount of time needed for verification. This improvement appears to be closely related to the number of nodes that were generated during the verification task. In addition, a reduced number of nodes also decreases the amount of memory required to store the partition of the input set $\mathcal{B}$.

Note from Table \ref{tbl:averages_stds} that 6 timeouts occurred for property $\phi_2$: 4 for IOG and 2 for BE. We believe BE-based splits outperformed IOG-based splits in these instances because the metric in \eqref{eq:metric_for_split} encourages splits to generate partitions that lead towards regions of the input space where all ReLU nodes are either active or inactive (i.e. towards the interior of the polytopic regions $\mathcal{P}_i$ of the domain), which leads to finding examples faster. In addition, when comparing networks which were able to be verified by \textit{both} procedures, BE still outperformed IOG in the amount of time and nodes generated as shown in Table \ref{tbl:comparison}. While BE-splits did under perform when verifying property $\phi_8$, it is important to note that this property is only verified against a single network. In constrast, property $\phi_2$ was tested against 30 different networks. In the Appendix we include information on property $\phi_2$ including timeouts.

\section{Conclusion}
\label{sec:conclusion}

In safety-critical application, such as self-driving, it is important to be able to verify the behavior learned by deep neural networks.
In this paper, we introduced a new technique for verification of ReLU NNs, that relies on splitting the input set.
Previous work leverages information of the gradient between inputs and outputs to decide how splits should be undertaken in order to verify a property of interest. 
In this work, we showed that by using shadow prices, a metric representing constraint sensitivity, and estimates on the bounds, we can substantially improve the amount of memory and time required to solve verification tasks. 
We test our proposed approach on the ACAS benchmark and provide a side-by-side comparison between both splitting procedures. 
Our results show substantial improvements both in the amount of memory and time required to solve these benchmarks.
Future work will focus on extending the verification to more general classes of NNs, as well as input sets with arbitrary topology.

% Acknowledgements should only appear in the accepted version.
% \section*{Acknowledgements}

% \textbf{Do not} include acknowledgements in the initial version of
% the paper submitted for blind review.

% If a paper is accepted, the final camera-ready version can (and
% probably should) include acknowledgements. In this case, please
% place such acknowledgements in an unnumbered section at the
% end of the paper. Typically, this will include thanks to reviewers
% who gave useful comments, to colleagues who contributed to the ideas,
% and to funding agencies and corporate sponsors that provided financial
% support.

\bibliography{main.bib}
\bibliographystyle{icml2019}

%%%%%%%%%%%%%%%%%%%%%%%%%%%%%%%%%%%%%%%%%%%%%%%%%%%%%%%%%%%%%%%%%%%%%%%%%%%%%%%
%%%%%%%%%%%%%%%%%%%%%%%%%%%%%%%%%%%%%%%%%%%%%%%%%%%%%%%%%%%%%%%%%%%%%%%%%%%%%%%
% DELETE THIS PART. DO NOT PLACE CONTENT AFTER THE REFERENCES!
%%%%%%%%%%%%%%%%%%%%%%%%%%%%%%%%%%%%%%%%%%%%%%%%%%%%%%%%%%%%%%%%%%%%%%%%%%%%%%%
%%%%%%%%%%%%%%%%%%%%%%%%%%%%%%%%%%%%%%%%%%%%%%%%%%%%%%%%%%%%%%%%%%%%%%%%%%%%%%%
\appendix
\clearpage
\section{Appendix}

\subsection{Property $\phi_2$ including timeouts}

Here we include the histogram of property $\phi_2$ with timeouts.

\begin{figure}[h]
\centering
\includegraphics[width=.8\columnwidth]{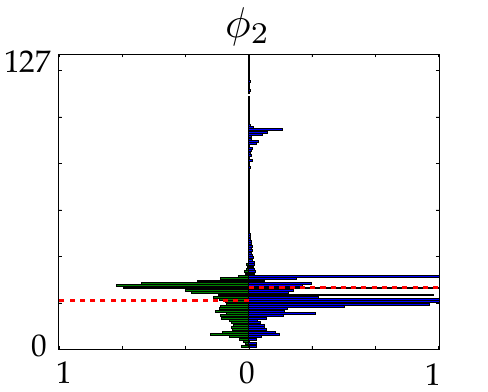} %width=\columnwidth,
\caption{Histogram of branch lengths for property 2 including the timeouts.}
\label{fig:extra}
\end{figure}

The IOG search depth was $26.59\pm16.48$ with $17619$ nodes generated. The BE search depth was $21.02\pm7.82$ with $6927$ nodes generated. The ratio of times $t_{BE}/t_{IOG}$ is $0.56$.

\subsection{ACAS Benchmark}

We now list the 10 properties contained in the Airborne Collision Avoidance System benchmark.

\paragraph{Property $\phi_1$:}
\begin{itemize}[noitemsep, nolistsep, leftmargin=*]
\item \textbf{Description:}
  If the intruder is distant and is
  significantly slower than the ownship, the score of a COC advisory will
  always be below a certain fixed threshold.
\item \textbf{Tested on:} all 45 networks.
\item \textbf{Input constraints:}
  $\rho\geq 55947.691$,
  $v_\text{own}\geq 1145$,
  $v_\text{int}\leq 60$.
\item \textbf{Desired output property:} the score for COC is at most $1500$.
\end{itemize}

\paragraph{Property $\phi_2$:}
\begin{itemize}[noitemsep, nolistsep, leftmargin=*]
\item \textbf{Description:}
  If the intruder is distant and is
  significantly slower than the ownship, the score of a COC advisory will
  never be maximal.
\item \textbf{Tested on:} $N_{x,y}$ for all $x\geq 2$ and for all $y$, except $N_{4,2}$ and $N_{5,3}$.
\item \textbf{Input constraints:}
  $\rho\geq 55947.691$,
  $v_\text{own}\geq 1145$,
  $v_\text{int}\leq 60$.
\item \textbf{Desired output property:} the score for COC is not the maximal score.
\end{itemize}

\paragraph{Property $\phi_3$:}
\begin{itemize}[noitemsep, nolistsep, leftmargin=*]
\item \textbf{Description:}
  If the intruder is directly ahead and is moving towards the ownship,
  the score for COC will not be minimal.
\item \textbf{Tested on:} all networks except $N_{1,7}$, $N_{1,8}$, and $N_{1,9}$.
\item \textbf{Input constraints:}  
    $1500 \leq \rho \leq 1800$,
    $-0.06 \leq \theta \leq 0.06$,
    $\psi \geq 3.10$,
    $v_\text{own}\geq 980$,
    $v_\text{int}\geq 960$.
\item \textbf{Desired output property:} the score for COC is not the minimal score.
\end{itemize}

\paragraph{Property $\phi_4$:}
\begin{itemize}[noitemsep, nolistsep, leftmargin=*]
\item \textbf{Description:}
  If the intruder is directly ahead and is moving away from the
  ownship but at a lower speed than that of the ownship,
  the score for COC will not be minimal.
\item \textbf{Tested on:} all networks except $N_{1,7}$, $N_{1,8}$, and $N_{1,9}$.
\item \textbf{Input constraints:}  
    $1500 \leq \rho \leq 1800$,
    $-0.06 \leq \theta \leq 0.06$,
    $\psi = 0$,
    $v_\text{own}\geq 1000$,
    $700 \leq v_\text{int}\leq 800$.
\item \textbf{Desired output property:} the score for COC is not the minimal score.
\end{itemize}

\paragraph{Property $\phi_5$:}
\begin{itemize}[noitemsep, nolistsep, leftmargin=*]
\item \textbf{Description:}
  If the intruder is near and approaching from the
  left, the network advises ``strong right''.
\item \textbf{Tested on:} $N_{1,1}$.
\item \textbf{Input constraints:}
    $250 \leq \rho \leq 400$,
    $0.2 \leq \theta \leq 0.4$,
    $-3.141592 \leq \psi \leq -3.141592 + 0.005$,
    $100 \leq v_\text{own}\leq 400$,
    $0 \leq v_\text{int}\leq 400$.
\item \textbf{Desired output property:} the score for ``strong right'' is the minimal score.
\end{itemize}

\paragraph{Property $\phi_6$:}
\begin{itemize}[noitemsep, nolistsep, leftmargin=*]
\item \textbf{Description:}
  If the intruder is sufficiently far away,
  the network advises COC.
\item \textbf{Tested on:} $N_{1,1}$.
\item \textbf{Input constraints:}
    $12000 \leq \rho \leq 62000$,
    $(0.7 \leq \theta \leq 3.141592)
    \vee
    (-3.141592 \leq \theta \leq -0.7)$,
    $-3.141592 \leq \psi \leq -3.141592 + 0.005$,
    $100 \leq v_\text{own}\leq 1200$,
    $0 \leq v_\text{int}\leq 1200$.
\item \textbf{Desired output property:} the score for COC is the minimal score.
\end{itemize}

\paragraph{Property $\phi_7$:}
\begin{itemize}[noitemsep, nolistsep, leftmargin=*]
\item \textbf{Description:}
  If vertical separation is large,
  the network will never advise a strong turn.
\item \textbf{Tested on:} $N_{1,9}$.
\item \textbf{Input constraints:}
    $0 \leq \rho \leq 60760$,
    $-3.141592 \leq \theta \leq 3.141592$,
    $-3.141592 \leq \psi \leq 3.141592$,
    $100 \leq v_\text{own}\leq 1200$,
    $0 \leq v_\text{int}\leq 1200$.
\item \textbf{Desired output property:} the scores for ``strong right'' and
  ``strong left'' are never the minimal scores.
\end{itemize}

\paragraph{Property $\phi_8$:}
\begin{itemize}[noitemsep, nolistsep, leftmargin=*]
\item \textbf{Description:}
 For a large vertical separation and a previous
 ``weak left'' advisory, the network will either output COC or
 continue advising ``weak left''.
\item \textbf{Tested on:} $N_{2,9}$.
\item \textbf{Input constraints:}
    $0 \leq \rho \leq 60760$,
    $-3.141592 \leq \theta \leq -0.75\cdot 3.141592$,
    $-0.1 \leq \psi \leq 0.1$,
    $600 \leq v_\text{own}\leq 1200$,
    $600 \leq v_\text{int}\leq 1200$.
\item
Desired output property: the score for ``weak left'' is minimal or the
score for COC is minimal.
\end{itemize}

\paragraph{Property $\phi_9$:}
\begin{itemize}[noitemsep, nolistsep, leftmargin=*]
\item \textbf{Description:}
Even if the previous advisory was ``weak right'',
the presence of a nearby intruder will cause the network to output a
 ``strong left'' advisory instead.
\item \textbf{Tested on:} $N_{3,3}$.
\item \textbf{Input constraints:}
    $2000 \leq \rho \leq 7000$,
    $-0.4 \leq \theta \leq -0.14$,
    $-3.141592 \leq \psi \leq -3.141592+0.01$,
    $100 \leq v_\text{own}\leq 150$,
    $0 \leq v_\text{int}\leq 150$.
\item \textbf{Desired output property:} the score for ``strong left'' is minimal.
\end{itemize}

\paragraph{Property $\phi_{10}$:}
\begin{itemize}[noitemsep, nolistsep, leftmargin=*]
\item \textbf{Description:}
For a far away intruder, the network advises COC.
\item \textbf{Tested on:} $N_{4,5}$.
\item \textbf{Input constraints:}
    $36000 \leq \rho \leq 60760$,
    $0.7 \leq \theta \leq 3.141592$,
    $-3.141592 \leq \psi \leq -3.141592+0.01$,
    $900 \leq v_\text{own}\leq 1200$,
    $600 \leq v_\text{int}\leq 1200$.
\item \textbf{Desired output property:} the score for COC is minimal.
\end{itemize}

%%%%%%%%%%%%%%%%%%%%%%%%%%%%%%%%%%%%%%%%%%%%%%%%%%%%%%%%%%%%%%%%%%%%%%%%%%%%%%%
%%%%%%%%%%%%%%%%%%%%%%%%%%%%%%%%%%%%%%%%%%%%%%%%%%%%%%%%%%%%%%%%%%%%%%%%%%%%%%%

\end{document}